\documentclass{article}
\usepackage{spconf,amsmath,graphicx}
\usepackage{xcolor}

\usepackage{amssymb}
\usepackage{stackengine}
\usepackage{scalerel}
\usepackage{multirow}

\title{Deep Neural Network based Cough Detection using Bed-mounted Accelerometer Measurements}
\name{Madhurananda Pahar$^1$$^\dagger$\sthanks{We would like to thank the South African Centre for High Performance Computing (CHPC) for providing computational resources on their Lengau cluster for this research.}, Igor Miranda$^1$$^\oplus$, Andreas Diacon$^2$$^\dagger$, Thomas Niesler$^1$$^\dagger$}
 
\address{$^1$Department of Electrical and Electronic Engineering,  University of Stellenbosch, South Africa\\
	$^2$TASK Applied Science, Cape Town, South Africa\\
	$^\dagger$\{mpahar, ahd, trn\}@sun.ac.za; $^\oplus$igordantas@ufrb.edu.br}

\begin{document}

\textbf{Copyright 2021 IEEE. Published in ICASSP 2021 - 2021 IEEE International Conference on Acoustics, Speech and Signal Processing (ICASSP), scheduled for 6-11 June 2021 in Toronto, Ontario, Canada. Personal use of this material is permitted. However, permission to reprint/republish this material for advertising or promotional purposes or for creating new collective works for resale or redistribution to servers or lists, or to reuse any copyrighted component of this work in other works, must be obtained from the IEEE. Contact: Manager, Copyrights and Permissions / IEEE Service Center / 445 Hoes Lane / P.O. Box 1331 / Piscataway, NJ 08855-1331, USA. Telephone: + Intl. 908-562-3966.}

\maketitle

\begin{abstract}
We have performed cough detection based on measurements from an accelerometer attached to the patient's bed.    
This form of monitoring is less intrusive than body-attached accelerometer sensors, and sidesteps privacy concerns encountered when using audio for cough detection.    
For our experiments, we have compiled a manually-annotated dataset containing the acceleration signals of approximately 6000 cough and 68000 non-cough events from 14 adult male patients in a tuberculosis clinic. 
As classifiers, we have considered convolutional neural networks (CNN), long-short-term-memory (LSTM) networks, and a residual neural network (Resnet50). 
We find that all classifiers are able to distinguish between the acceleration signals due to coughing and those due to other activities including sneezing, throat-clearing and movement in the bed with high accuracy.    
The Resnet50 performs the best, achieving an area under the ROC curve (AUC) exceeding 0.98 in cross-validation experiments.
We conclude that high-accuracy cough monitoring based only on measurements from the accelerometer in a consumer smartphone is possible.    
Since the need to gather audio is avoided and therefore privacy is inherently protected, and since the accelerometer is attached to the bed and not worn, this form of monitoring may represent a more convenient and readily accepted method of long-term patient cough monitoring.
\end{abstract}


\section{Introduction}

Coughing is the forceful expulsion of air to clear up the airway and a common symptom of respiratory disease~\cite{korpavs1996analysis}. 
It can be distinctive in nature and is an important indicator used by physicians for clinical diagnosis and heath monitoring in more than 100 respiratory diseases~\cite{knocikova2008wavelet}, including tuberculosis (TB) \cite{botha2018detection}, asthma \cite{al2013signal} and pertussis~\cite{pramono2016cough}.  

Automatic cough detection and classification is possible by applying machine learning algorithms on extracted features from cough sounds~\cite{miranda2019comparative}.
It has also been shown to be possible when using the signals from an accelerometer  
placed on the patient's body~\cite{mohammadi2019automatic}. 
Since the accelerometer is insensitive to environmental and background noise, it can be used in conjunction with other sensors such as microphones, ECG and thermistors~\cite{munyard1994new}. 

A cough monitoring system using a contact microphone and an accelerometer attached to the participant's suprasternal (jugular) notch has been considered by~\cite{pavesi2001application}.
This system allows participants to move around within their homes while the cough audio and vibration is recorded. 
In related work, an ambulatory cough monitoring system using an accelerometer attached to the skin of the participant's suprasternal notch using a bioclusive transparent dressing was developed in~\cite{paul2006evaluation}.
Here the recorded signal is transmitted to a receiver carried in a pocket or attached to a belt.  

Throat-mounted accelerometers have been used successfully to detect coughing in~\cite{coyle2010systems} and in~\cite{fan2014cough}, and an accelerometer placed at the laryngeal prominence (Adam's apple) in~\cite{mohammadi2019automatic}.
Two accelerometers, one placed on the abdomen and the second on a belt wrapped at dorsal region, have been used to measure cough rate in the research carried out by~\cite{chan2014systems}. 
Abdominal placement (between the navel and sternal notch) of the accelerometer was also investigated in~\cite{hirai2015new}, and it was applied to patients who were children.
Finally, multiple sensors, including ECG, thermistor, chest belt, accelerometer and audio microphones were used for cough detection in~\cite{drugman2013objective}. 

Attaching an accelerometer to the patient's body is however inconvenient and intrusive.
Thus, we propose the monitoring of coughing based on the signals obtained from the accelerometer inbuilt in an inexpensive consumer smartphone firmly attached to the patient's bed, as shown in Figure~\ref{fig:recorder-position}, thereby eliminating the need to wear a measuring equipment.
We have trained and evaluated deep neural network (DNN) classifiers such as
convolutional neural networks (CNN), long-short-term-memory (LSTM) networks, and a residual neural network (Resnet50) \cite{he2016deep} architecture
using leave-one-out cross-validation on a dataset, prepared for this purpose, which consists of cough and non-cough events such as sneezing, throat-clearing and getting in and out of the bed. 
The Resnet50 produces the highest AUC of 0.9888 after 50 epochs with corresponding accuracy 96.71\% and sensitivity 99\% for 32 sample (320 msec) long frames and grouping them in 10 segments. 
This shows that it is possible to discriminate between cough events and other non-cough events by using state-of-the-art classifiers such as a Resnet50 architecture; where the accelerometer is no longer attached to the patient's body, rather built inside an inexpensive consumer smartphone attached to the headboard of the patient's bed.

\section{Dataset Preparation}

Data collection was performed at a small 24h TB clinic near Cape Town, South Africa, which can accommodate approximately 10 staff and 30 patients. 
Each ward has four beds and can thus accommodate up to four patients at one time. 
The overall motivation of our work is to develop a practical method of automatic cough monitoring for the patients in this clinic, in order to assist with the monitoring of recovery.

\begin{figure}[h!]
	\centerline{\includegraphics[width=0.5\textwidth]{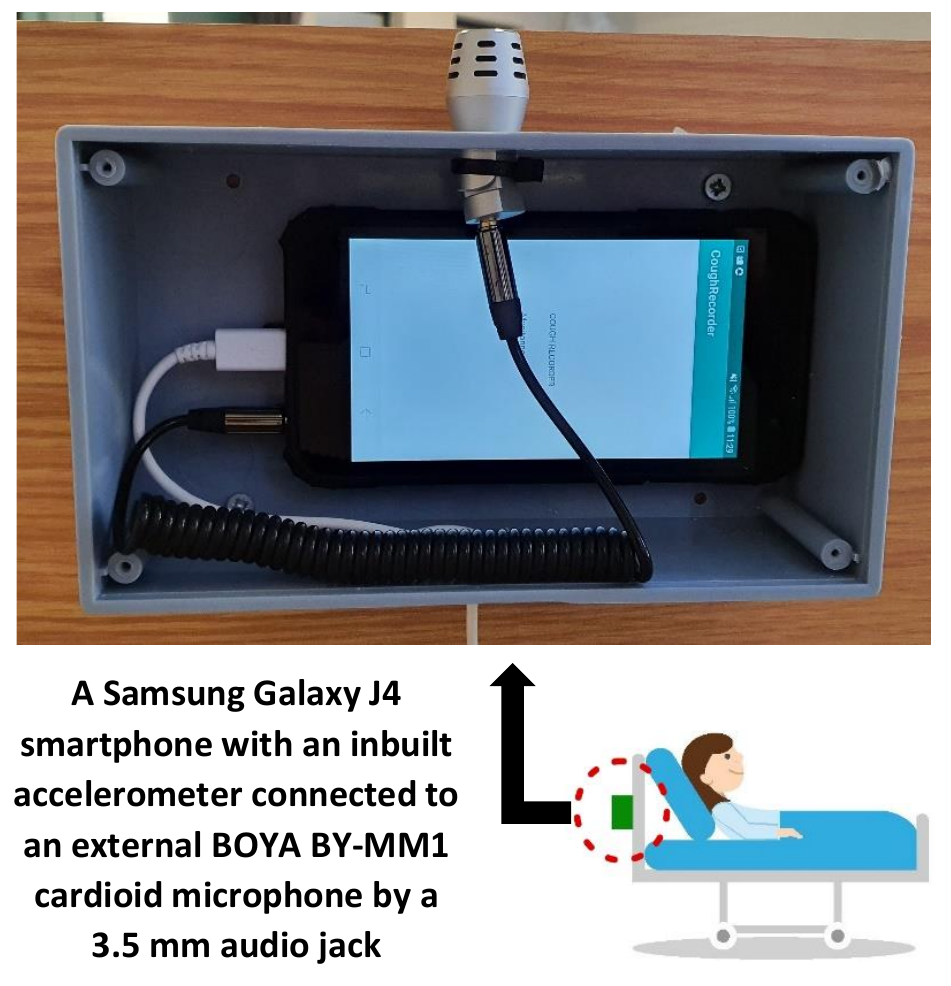}}
	\caption{\textbf{Recording Process:} A plastic enclosure housing an inexpensive smartphone running data gathering software is attached behind the headboard of each bed. The audio signal from the external microphone is also monitored, but only used for the purpose of annotation of the acceleration signal. }
	\label{fig:recorder-position}
\end{figure}

The recording setup is shown in Figure \ref{fig:recorder-position}. 
An enclosure housing an inexpensive consumer smartphone is firmly attached to the back of the headboard of each bed in a ward.
A data gathering Android smartphone application, developed for this study, continuously monitors the accelerometer and audio signals from an external microphone (also shown in Figure \ref{fig:recorder-position}). 
The 3-axis accelerometer has a sampling frequency of $100$ Hz and only magnitudes were recorded.
This energy-threshold-based detection for both audio and acceleration signals results in a large volume of data being captured. 
In addition, continuous video recordings were made using ceiling-mounted cameras in order to assist with data annotation. 

\begin{figure}[!h]
	\centerline{\includegraphics[width=0.5\textwidth]{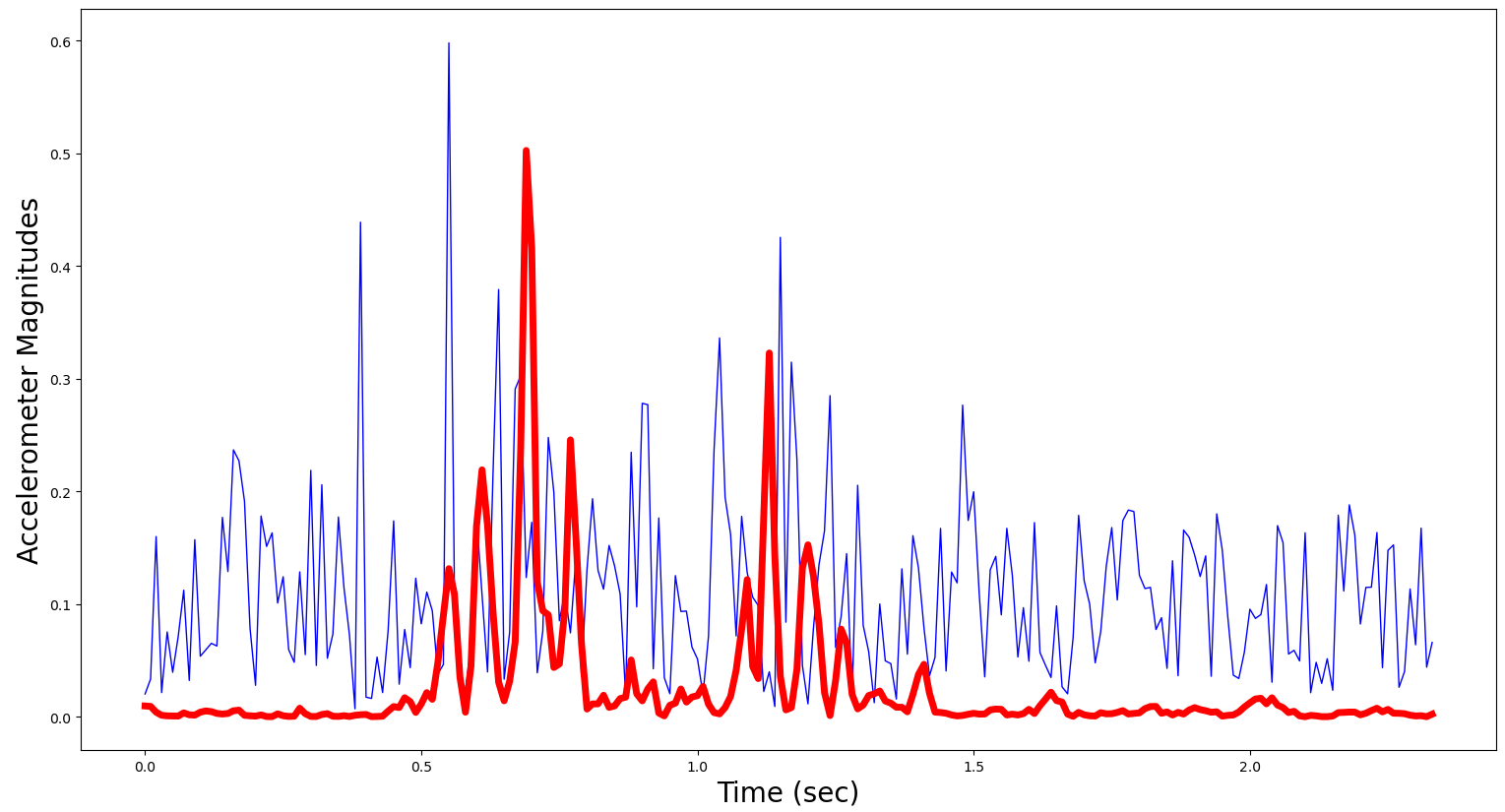}}
	\caption{\textbf{Example accelerometer magnitudes} for a cough event (red) and a non-cough event (blue). In this case, the non-cough event was the patient moving in the bed. 
	}
	\label{fig:cough-Ncough-acc}
\end{figure}

This work considers automatic classification of the acceleration signals.
The audio signals and video recordings were used only during the manual annotation process in order to unambiguously confirm the presence or absence of a cough.
We define an `event' to be any detected accelerometer or audio activity.
An example of the accelerometer magnitudes captured for a cough and for a non-cough event are shown in Figure~\ref{fig:cough-Ncough-acc}. 
Annotation was performed using the multimedia software tool ELAN \cite{wittenburg2006elan}, which allowed easy consolidation of the audio and video for accurate manual labelling.

Our final dataset, summarised in Table~\ref{table:Ground-Truth-Dataset-Summary}, contains approximately 6000 coughs and 68000 non-coughs from 14 adult male patients totalling 3.16 and 32.20 hours of data respectively. No other information regarding patients are recorded due to ethical constraints. 
This dataset was used to train and evaluate the classifiers within a cross-validation framework.

\begin{table}[h!]
	\footnotesize
	\caption{\textbf{Ground Truth Dataset Summary:} `\uppercase{Patients}': list of the patients; `\uppercase{Coughs}': number of confirmed cough events; `\uppercase{Non Coughs}': number of confirmed events that are not coughs; `\uppercase{Cough time}': total amount of time (in sec) for cough events; `\uppercase{Non-cough time}': total amount of time (in sec) for non-cough events. } 
	\centering 
	\begin{center}
		\begin{tabular}{ c c c c c }
			\hline
			\hline
			 {\multirow{2}{*}{\textbf{\uppercase{Patients}}}} & {\multirow{2}{*}{\textbf{\uppercase{Coughs}}}} & \textbf{\uppercase{Non}} & \textbf{\uppercase{Cough}} & \textbf{\uppercase{Non-}} \\
			 &  & \textbf{\uppercase{coughs}} & \textbf{\uppercase{time}} & \textbf{\uppercase{cough time}} \\
			\hline 
			\hline
			Patient 1 & 88 & 973 & 169.16 & 1660.67 \\
			
			\hline
			Patient 2 & 63 & 1111 & 117.67 & 1891.92 \\
			
			\hline
			Patient 3 & 469 & 11025 & 893.91 & 18797.32 \\
			
			\hline
			Patient 4 & 109 & 9151 & 204.06 & 15596.71 \\
			
			\hline
			Patient 5 & 97 & 7826 & 188.26 & 13344.98 \\
			
			\hline
			Patient 6 & 192 & 12437 & 360.72 & 21197.35 \\
			
			\hline
			Patient 7 & 436 & 14053 & 825.23 & 23953.15 \\
			
			\hline
			Patient 8 & 368 & 2977 & 702.05 & 5077.89 \\
			
			\hline
			Patient 9 & 2816 & 3856 & 5345.27 & 6569.32 \\
			
			\hline
			Patient 10 & 649 & 2579 & 1236.84 & 4400.42 \\
			
			\hline
			Patient 11 & 205 & 527 & 391.42 & 901.38 \\
			
			\hline
			Patient 12 & 213 & 323 & 402.61 & 547.62 \\
			
			\hline
			Patient 13 & 213 & 712 & 401.61 & 1211.75 \\
			
			\hline
			Patient 14 & 82 & 455 & 158.77 & 777.64 \\
			
			\hline
			\textbf{TOTAL} & \textbf{6000} & \textbf{68005} & \textbf{11397.6} & \textbf{115928.12} \\
			
			\hline
			\hline
		\end{tabular}
	\end{center}
	\label{table:Ground-Truth-Dataset-Summary}
\end{table}

Table~\ref{table:Ground-Truth-Dataset-Summary} shows that coughs are underrepresented in our dataset.
To compensate for this imbalance, which can detrimentally affect machine learning~\cite{van2007experimental,krawczyk2016learning}, we have applied SMOTE data balancing during training \cite{chawla2002smote,lemaitre2017imbalanced}. 
This technique oversamples the minor class by generating synthetic samples (instead of for example random oversampling).
SMOTE has previously been successfully applied to cough detection and classification based on audio recordings \cite{windmon2018tussiswatch}.

\section{Feature Extraction}

Power spectra~\cite{liang2013application}, 
root mean square value,
kurtosis, moving average and crest factor are extracted from the acceleration signals as features for classification. 
The frame length ($\Psi$) and number of segments ($C$) have been used as the feature extraction hyperparameters, shown in Table \ref{table:feat-hyper-parameter} and \ref{table:classifier-summary-hyper-avg}. 
The input feature matrix, shown in Figure \ref{fig:CNN-fig} and \ref{fig:RNN-fig}, has the dimension of ($C$, $\frac{\Psi}{2}+5$) and power spectra have $(\frac{\Psi}{2}+1)$ coefficients. 

\begin{table}[h]
	\footnotesize
	\caption{\textbf{Feature extraction hyperparameters.} Frame lengths (16, 32, 64 samples i.e. 160, 320 and 640 msec long) overlap in such a way that the number of segments (5 and 10) are the same for all events in the dataset.} 
	\centering 
	\begin{center}
		\begin{tabular}{ c | c | c }
			\hline
			\textbf{Hyperparameter} & \textbf{Description} & \textbf{Range} \\
			\hline
			\hline
			{\multirow{2}{*}{Frame length ($\Psi$)}} & Size of frames in samples & $2^k$ where \\
			 & in which cough is segmented &  $k=4, 5, 6$\\
			\hline
			{\multirow{2}{*}{No. of Segments ($C$)}} & Number of segments in & {\multirow{2}{*}{5, 10}} \\
			 & which frames were grouped &  \\
			\hline
			\hline
		\end{tabular}
	\end{center}
	\label{table:feat-hyper-parameter}
\end{table}

Frame length ($\Psi$) used to extract features from acceleration signal is shorter than those generally used to extract features from audio \cite{takahashi2016acoustic}, because 
the accelerometer in the smartphone (shown in Figure \ref{fig:recorder-position}) has a lower sampling rate of 100 Hz and
longer frames lead to deteriorated performance as the signal properties can no longer be assumed to be stationary.

\section{Classifier training}

Our dataset contains 14 patients and a leave-one-out cross-validation scheme \cite{sammut2010leave} has been used to train and evaluate our three DNN classifiers: CNN, LSTM and Resnet50. 

\begin{figure}
	\centerline{\includegraphics[width=0.5\textwidth]{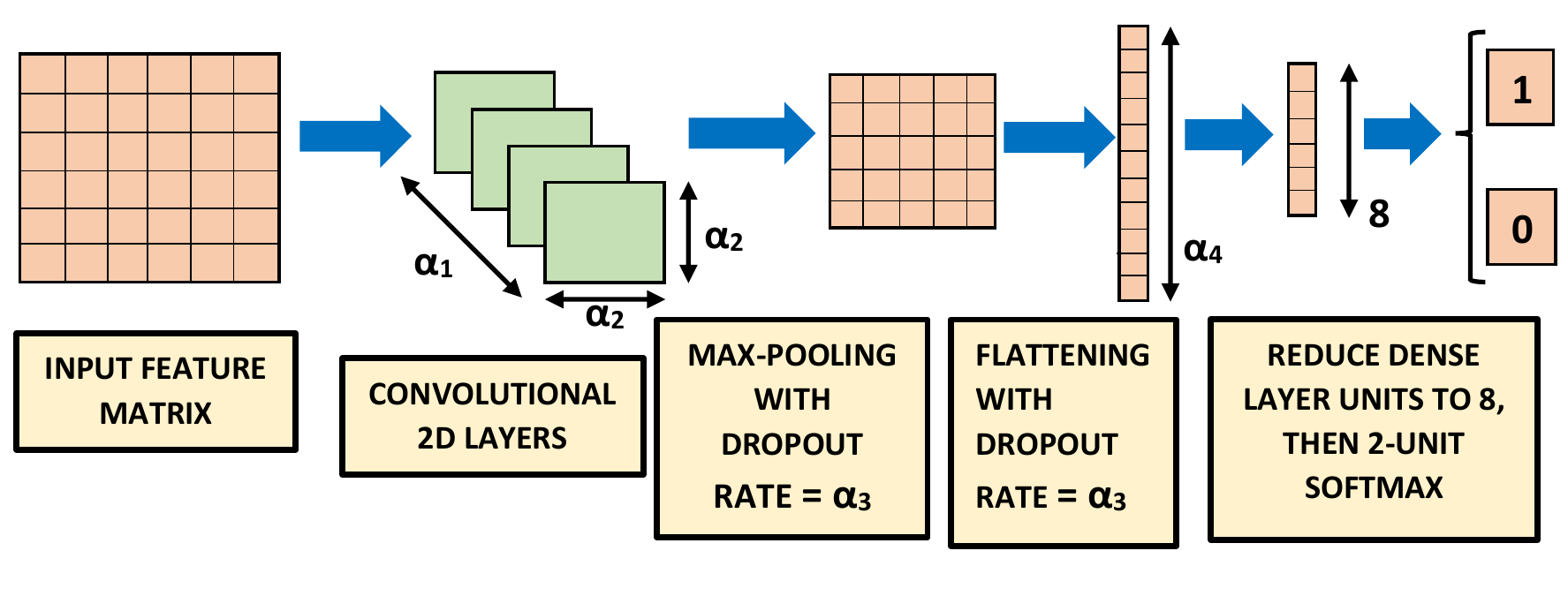}}
	\caption{\textbf{CNN Classifier}, trained and evaluated using leave-one-out cross-validation \cite{sammut2010leave}, produces results shown in Table \ref{table:classifier-summary-hyper-avg} for feature extraction hyperparameters shown in Table \ref{table:feat-hyper-parameter}.}
	\label{fig:CNN-fig}
\end{figure}

The CNN classifier, shown in Figure \ref{fig:CNN-fig}, has been set up with $\alpha_1$ 2D convolutional layers with kernel size $\alpha_2$ and rectified linear units as activation functions. A dropout rate $\alpha_3$ has been applied along with max-pooling, followed by 
$\alpha_4$ dense layers with rectified linear units as activation functions, followed by another 8 dense layers, also with rectified linear units as activation functions. 

\begin{figure}
	\centerline{\includegraphics[width=0.5\textwidth]{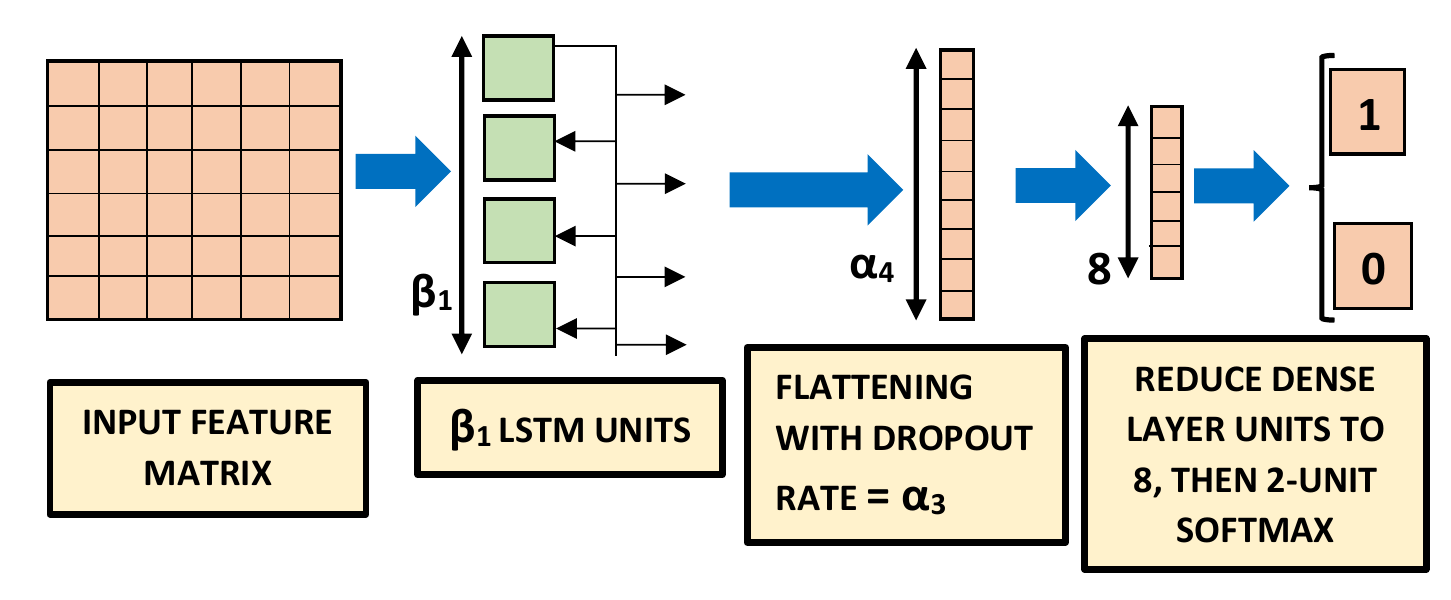}}
	\caption{\textbf{LSTM classifier}, trained and evaluated using leave-one-out cross-validation \cite{sammut2010leave}, produces results shown in Table \ref{table:classifier-summary-hyper-avg} for feature extraction hyperparameters in Table \ref{table:feat-hyper-parameter}.}
	\label{fig:RNN-fig}
\end{figure}

The LSTM classifier, shown in Figure \ref{fig:RNN-fig}, has been set up with $\beta_1$ LSTM units with rectified linear units as activation functions and a dropout rate $\alpha_3$. 
Then $\alpha_4$ dense layers have been applied with rectified linear units as activation functions, followed by another 8 dense layers also with rectified linear units as activation functions. 

For both LSTM and CNN classifiers, a final softmax function produces one output for a cough event (i.e. 1) and the other for a non-cough event (i.e. 0), as shown in Figure \ref{fig:CNN-fig} and \ref{fig:RNN-fig}. 
Features are fed into these two classifiers in batch size of $\xi_1$ for $\xi_2$ number of epochs. 

The residual network (Resnet) architecture we trained and evaluated has 50 layers and has been found to deliver the state-of-the-art performance in image recognition.
We have replicated the 50-layer architecture used in Table 1 of \cite{he2016deep} in our experiments.
Table~\ref{table:class-hyper-parameter} lists the classifier hyperparameters that were optimised during leave-one-out cross-validation.

\begin{table}[h]
	\footnotesize
	\caption{\textbf{CNN \& LSTM classifier hyperparameters}, optimised using leave-one-out cross-validation and shown in Figure \ref{fig:CNN-fig} and \ref{fig:RNN-fig}. } 
	\centering 
	\begin{center}
		\begin{tabular}{ c | c | c  }
			\hline
			\textbf{Hyperparameters} & \textbf{Classifier} & \textbf{Range} \\
			\hline
			\hline
			Batch Size ($\xi_1$) & CNN \& LSTM & $2^k$ where $k=6, 7, 8$\\
			\hline
			No. of epochs ($\xi_2$) & CNN \& LSTM & 10 to 200 in steps of 20 \\
			\hline
			No. of Conv filters ($\alpha_1$) & CNN & $3 \times 2^k$ where $k=3, 4, 5$ \\
			\hline
			kernel size ($\alpha_2$) & CNN & 2 and 3 \\
			\hline
			Dropout rate ($\alpha_3$) & CNN \& LSTM & 0.1 to 0.5 in steps of 0.2 \\
			\hline
			Dense layer size ($\alpha_4$) & CNN \& LSTM & $2^k$ where $k=4, 5$ \\
			\hline
			LSTM units ($\beta_1$) & LSTM & $2^k$ where $k=6, 7, 8$ \\
			\hline
			Learning rate ($\beta_2$) & LSTM & $10^k$ where $k=-2,-3,-4$ \\
			\hline
			\hline
		\end{tabular}
	\end{center}
	\label{table:class-hyper-parameter}
\end{table}

\begin{figure}
	\centerline{\includegraphics[width=0.5\textwidth]{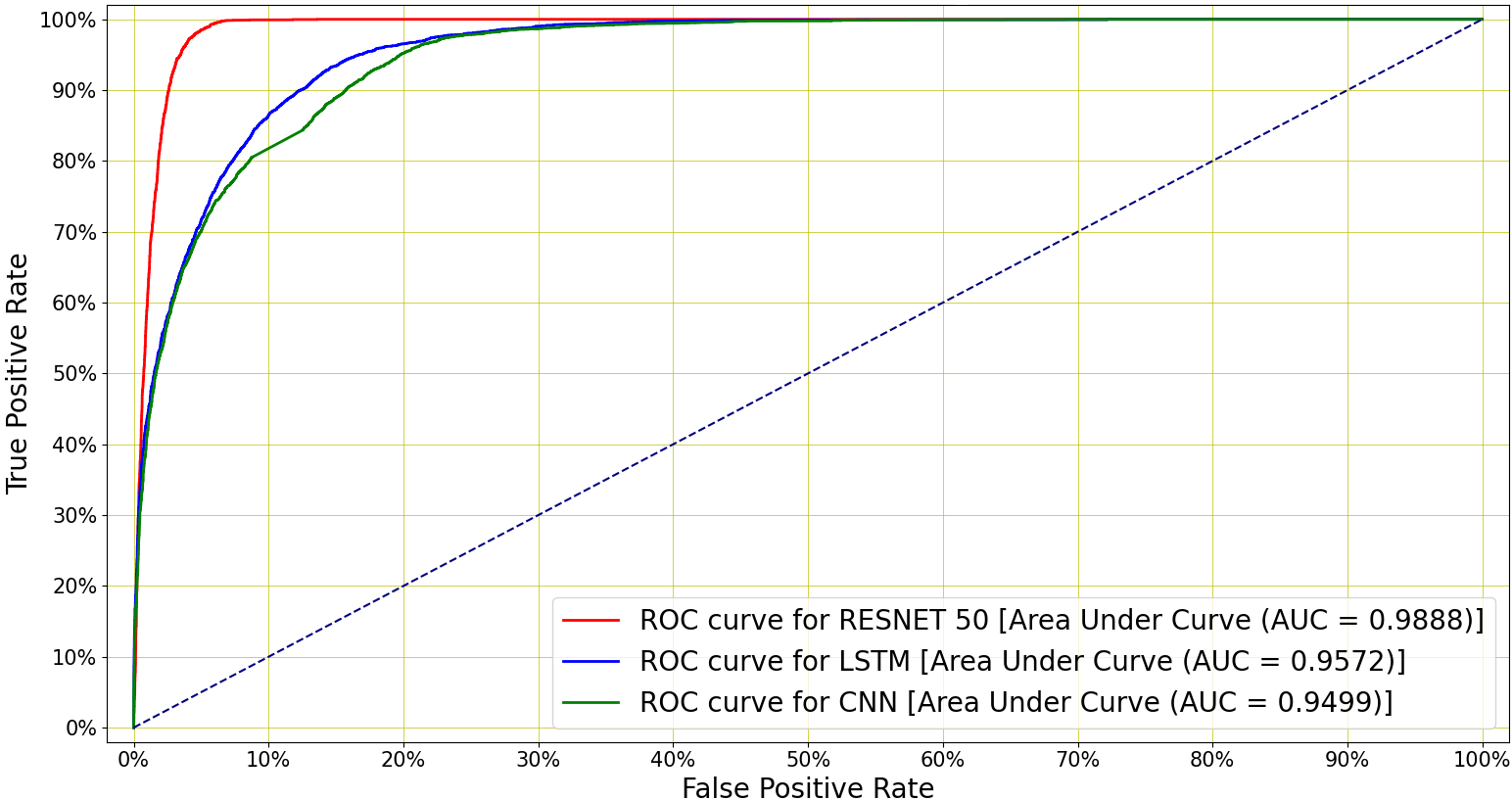}}
	\caption{\textbf{Mean ROC curves for cough detection}, for the best performing DNN classifiers in Table \ref{table:classifier-summary-hyper-avg}. AUC values are averaged over 14 leave-one-patient-out cross-validation folds during hyperparameter optimisation. Resnet50 outperforms the LSTM and CNN over a wide range of operating points and has achieved the highest accuracy of 96.71\%.}
	\label{fig:mean-ROC}
\end{figure}

\section{Results}

Table~\ref{table:classifier-summary-hyper-avg} lists the performance achieved by our three DNN classifiers for the hyperparameters mentioned in Table \ref{table:feat-hyper-parameter}.
These results are averages over the 14 leave-one-patient-out testing partitions during hyperparameter optimisation. 

Table~\ref{table:classifier-summary-hyper-avg} shows that the best-performing CNN uses 64 samples (640 msec) long frames and 10 number of segments to achieve an AUC of 0.9499. 
The optimal LSTM classifier achieves the slightly higher AUC of 0.9572 when using a frame length of 32 samples (320 msec) and 10 number of segments.
However, the best performance is achieved by the Resnet50 architecture, with an AUC of 0.9888 after 50 epochs from 32 samples (320 msec) long frames and 10 number of segments. 
Figure~\ref{fig:mean-ROC} shows the mean ROC curves for the optimal CNN, LSTM and Resnet50 configurations shown in Table \ref{table:classifier-summary-hyper-avg}, where the means were calculated over the 14 cross-validation folds. 
The Resnet50 classifier is superior to the other two classifiers over a wide range of operating points.

\begin{table}[h]
	\footnotesize
	\caption{\textbf{Leave-one-out cross-validation results for DNN classifiers.} The values are averaged over 14 cross-validation folds.} 
	\centering 
	\begin{center}
		\begin{tabular}{ c c c c c c c }
			\hline
			{\textbf{Frame}} & {\textbf{Seg}} & \textbf{Clas-} & \textbf{Mean} & \textbf{Mean} & \textbf{Mean} & \textbf{Mean} \\
			\textbf{($\Psi$)} & \textbf{($C$)} & \textbf{-sifier} & \textbf{Spec} & \textbf{Sens} & \textbf{Accuracy} & \textbf{AUC} \\
			\hline
			
			\hline
			16 & 5 & CNN & 83\% & 86\% & 84.55\% & 0.9243 \\
			\hline
			16 & 10 & CNN & 87\% & 84\% & 85.66\% & 0.9358 \\
			
			\hline
			32 & 5 & CNN & 76\% & 93\% & 84.47\% & 0.9272 \\
			\hline
			32 & 10 & CNN & 84\% & 86\% & 85.25\% & 0.9324 \\
			
			\hline
			64 & 5 & CNN & 85\% & 87\% & 86.31\% & 0.9339 \\
			\hline
			\textit{64} & \textit{10} & \textit{CNN} & \textit{91\%} & \textit{80\%} & \textit{85.82\%} & \textit{0.9499} \\			
			\hline
			
			\hline
			
			\hline
			16 & 5 & LSTM & 84\% & 91\% & 87.58\% & 0.9444 \\
			\hline
			16 & 10 & LSTM & 85\% & 92\% & 88.32\% & 0.9504 \\
			
			\hline
			32 & 5 & LSTM & 79\% & 95\% & 87.1\% & 0.9457 \\
			\hline
			\textit{32} & \textit{10} & \textit{LSTM} & \textit{86\%} & \textit{93\%} & \textit{89.21\%} & \textit{0.9572} \\
			
			\hline
			64 & 5 & LSTM & 84\% & 93\% & 88.68\% & 0.954 \\
			\hline
			64 & 10 & LSTM & 86\% & 89\% & 87.66\% & 0.9489 \\			
			\hline
			
			\hline
			
			\hline
			16 & 5 & Resnet50 & 93\% & 98\% & 95.43\% & 0.9802 \\
			\hline
			16 & 10 & Resnet50 & 94\% & 99\% & 96.35\% & 0.9812 \\
			
			\hline
			32 & 5 & Resnet50 & 94\% & 99\% & 96.54\% & 0.9810 \\
			\hline
			\textit{\textbf{32}} & \textit{\textbf{10}} & \textit{\textbf{Resnet50}} & \textit{\textbf{94\%}} & \textit{\textbf{99\%}} & \textit{\textbf{96.71\%}} & \textit{\textbf{0.9888}} \\
			
			\hline
			64 & 5 & Resnet50 & 94\% & 99\% & 96.35\% & 0.9854 \\
			\hline
			64 & 10 & Resnet50 & 95\% & 98\% & 96.46\% & 0.9884 \\			
			\hline

			\hline

		\end{tabular}
	\end{center}
	\label{table:classifier-summary-hyper-avg}
\end{table}

\section{Conclusion and Future Work}
A deep neural network based cough detector is able to accurately discriminate between the accelerometer measurements due to coughing and due to other movements as captured by a consumer smartphone attached to a patient's bed.
The best system, using the Resnet50 architecture, achieves an AUC of 0.9888.
These experimental results are based on a specially-compiled corpus of manually-annotated acceleration measurements, including approximately 6000 cough and 68000 non-cough events, gathered from 14 patients in a small TB clinic.
Although accelerometer-based detection of coughing has been considered before, it has always made use of sensors worn by the patient, which is in some respects intrusive and can be inconvenient.
We have shown that excellent discrimination is also possible when the sensor is attached to the patient's bed.
This presents a less intrusive method of cough monitoring, which can be of practical use in monitoring the recovery process of patients, for example in the clinic where the data was collected.
Acceleration-based monitoring also has important privacy advantages over the detection of cough sounds by audio- based monitoring,
which often raises privacy concerns, and we have found patients to be uncomfortable in the presence of audio-based monitoring equipment.
Accelerometer-based monitoring, using a bed-mounted inexpensive consumer smartphone, represents a more discreet and also cost-effective alternative.

In ongoing work, we are attempting to optimise some of the Resnet50 metaparameters and also to incorporate audio \cite{pahar_coding_2020} along with accelerometer measurements with a view to further improvement on cough detection results.

\bibliographystyle{IEEEbib}
\small{
	\bibliography{reference}
}

\end{document}